# Guidelines and Annotation Framework for Arabic Author Profiling

Wajdi Zaghouani,[1] Anis Charfi[2]

[1] College of Humanities and Social Sciences, Hamad Bin Khalifa University, Qatar
[2] Information Systems Program, Carnegie Mellon University Qatar
E-Mail: wzaghouani@hbku.edu.qa, acharfi@qatar.cmu.edu

**Abstract**

In this paper, we present the annotation pipeline and the guidelines we wrote as part of an effort to create a large manually annotated Arabic author profiling dataset from various social media sources covering 16 Arabic countries and 11 dialectal regions. The target size of the annotated ARAP-Tweet corpus is more than 2.4 million words. We illustrate and summarize our general and dialect-specific guidelines for each of the dialectal regions selected. We also present the annotation framework and logistics. We control the annotation quality frequently by computing the inter-annotator agreement during the annotation process. Finally, we describe the issues encountered during the annotation phase, especially those related to the peculiarities of Arabic dialectal varieties as used in social media.

**Keywords:** Guidelines, Annotation, Corpus, Arabic, Social media

## 1. Introduction

Research on author profiling has always been constrained by the limited availability of training data. In fact, collecting textual data with the appropriate meta-data requires significant collection and annotation efforts. For every text, the characteristics of the author have to be known in order to successfully profile the author. Moreover, when the text is written in a dialectal variety such as the Arabic text used in social media, author profiling becomes even more challenging as it requires representative annotated datasets to be available for each dialectal variety.

Arabic dialects are historically related to the classical Arabic and they co-exist with the Modern Standard Arabic (MSA) in a diglossic relation. While standard Arabic has a clearly defined set of orthographic standards, the various Arabic dialects have no official orthographies and a given word could be written in multiple ways in different Arabic dialects as shown in Table 1.

This paper presents the guidelines and annotation work carried out within the Qatar National Research Fund (QNRF) research project on Arabic Author Profiling for Cyber-Security (ARAP)[1]. We used these guidelines in order to create resources and tools for 11 Arabic dialects (Zaghouani and Charfi 2018a; Zaghouani and Charfi 2018b). We collected our ARAP-Tweet corpus data from public Twitter accounts across various regions in the Arab world.

For the author profiling task, most of the currently available resources are for English and other European languages as described by (Celli et al., 2013). The dialectal Arabic resources are still lagging behind other languages when it comes to the availability of the required datasets (Rosso et al., 2018; Zaghouani, 2014).

To the best of our knowledge, there is no dialectal Arabic corpus available for the detection of age, gender, native language and dialectal variety. Having a large amount of annotated data remains the key to reliable results for tasks such as the author profiling. The lack of such resources motivated the creation of the resources presented in this paper.

Once we collected the dataset, we wrote the guidelines for the annotation of Tweets collected for their dialectal variety, their native language, the gender of the user and the age within three categories (under 25 years, 25 to 34 and 35 and above). Furthermore, we hired a team of experienced annotators and we designed an optimized annotation workflow. Moreover, we followed a consistent annotation evaluation protocol in order to validate our annotation protocol.

| Variety | Sentence |
|---|---|
| **English** | **When I went to the library** |
| **Standard Arabic** | عندما ذهبت إلى المكتبة <br> ʿindamā ḏahabtu ʾila l-maktabati |
| **Tunisian** | وقتلي مشيت للمكتبة <br> wăqtəlli mʃit l-əl-măktba |
| **Algerian** | ملي رحت للمكتبة <br> məlli raħt l-əl-măktaba |
| **Moroccan** | ملي مشيت للمكتبة <br> məlli mʃit lmăktaba |
| **Egyptian** | اما رحت المكتبة <br> amma roħt el-maktaba |
| **Lebanese** | لما رحت عالمكتبة <br> lamma reħit ʕal-mektebe |
| **Iraqi** | من رحت للمكتبة <br> min reħit lil-maktaba |
| **Qatari** | لمن رحت المكتبة <br> lamman ruħt el-maktaba |

Table 1: A sample sentence in seven Arabic Dialects

Overall, our corpus has the following features that distinguish it from other Arabic annotation projects:

• **Aim:** designed mainly as a resource for developing Author profiling tools.

---
[1] http://arap.qatar.cmu.edu/

- **Size:** 2,4 million words.
- **Text types:** Social Media from Twitter
- **Variety:** our data is from 16 Arabic countries representing 11 major Arabic regional dialects.

The remainder of this paper is organized as follows. In Section 2, we discuss related work. Then, we present our ARAP-Tweet corpus collected in Section 3. Section 4 describes our annotation guidelines whereas Section 5 explains our annotation logistics and workflow. Section 6 presents the evaluation of the annotation quality.

## 2. Related Work

We identified several efforts to create resources for some major Arabic dialects such as Egyptian and Levantine (Diab and Habash, 2007, Pasha et al., 2014, Habash et al., 2013). Within the context of the Qatar Arabic Language Bank (QALB) project, a large-scale annotated corpus of users' comments, the dialectal words were marked and replaced by their equivalent in standard Arabic (Zaghouani et al., 2014; Zaghouani et al., 2015; Zaghouani et al., 2016a.)

In the same context, Salloum and Habash (2013), Sajjad et al. (2013), Salloum and Habash (2013) and Sawaf (2010) used a translation of dialectal Arabic to Standard Arabic as a pivot to translate to English. Zbib et al. (2012) used crowdsourcing approaches to create some resources for machine translation of Arabic dialects.

Al-Sabbagh and Girju (2010) extracted various cues from the Internet to create a lexicon from Dialectal to Modern Standard Arabic. Chiang et al. (2006) built a parser for Dialectal Arabic using the training data from the standard Arabic Penn Treebank. Boujelbane et al. (2013) created a dictionary based on the relation between MSA and Tunisian Arabic.

For the regional dialects, some existing projects were related to dialect identification as mentioned in (Habash et al., 2008; Elfardy and Diab, 2013; Zaidan and Callison-Burch, 2013).

Furthermore, a Twitter dialectal Arabic corpus was created by Mubarak and Darwish (Mubarak and Darwish, 2014) covering four dialectal regions using geolocation information associated with Twitter data.

As the dialectal Arabic is widely used nowadays in most of the informal communication online across the various regions of the Arab world such as in chats, emails, forums and social media, several research efforts were initiated to create dialectal Arabic dedicated tools and resources. However, many of these efforts were disjointed and not coordinated and most of them have only focused on a limited number of dialects or regions that cannot represent the different regions of the Arab world. For instance, some of these resources are not fine-grained with only four major dialectal regions represented such as North Africa, Levant, Egypt, and the Gulf.

For the Arabic author profiling task, the data to be collected is expected to be representative of most of the Arabic dialects and for the moment such resources are not yet available. We found only two projects related to that topic by Abbasi and Chen (2005) and Estival et. al. (2008). The first work focuses on author identification in English and Arabic web forum messages to automatically detect extremist groups. The second work focuses on author profiling for English and Arabic e-mails.

Recently, Bouamor et al. (2018) and Habash et al. (2018) built MADAR and wrote dialectal Arabic unified guidelines to create dialectal Arabic corpus and lexicon covering dialects of various cities across the Arab world with a focus on a travel domain corpus.

For the first time, during the Author Profiling task at PAN 2017 (Rangel et al. 2017),[2] an Arabic task was presented to identify the gender and the dialect using a corpus of four Arabic dialects namely, the North African dialect, the Egyptian Arabic, the Levantine Arabic and the Gulf Arabic. For the resources cited above, the domain was limited in one case and the coverage was limited to only four countries out of 22 Arabic countries in another case.

In our project, we support the major dialects in the Arab world by covering 11 regions and 16 countries. Hence, our project will provide important contributions to Arabic Author profiling.

## 3. Corpus Description

In this section, we describe the corpus collection and data selection processes carried out to locate and crawl users for each dialect group. For practical reasons, we harvested our data from Twitter as it provides a powerful and free API for crawling and collecting data about public Twitter accounts and public Tweets.

Using the Twitter API and the TweePy[3] library for Python, we collected tweets that contained typical dialectal distinct words generally used by speakers of a given dialect. In other words, we searched for tweets that use dialect specific words and expressions, which allowed us to restrict the tweets to the selected region as much as possible. For example, the word كرهبة /karhba/ 'car' in Tunisian Arabic or the word زول /zo:l/ 'man' in Sudanese Arabic. The seed words for each region were created following a study to identify several seed words for each region. Furthermore, the annotators were trained to identify the cases where a given seed word was used in a profile from another region.

During a six weeks period, we sampled our list of user profiles according to this method. Once we had the initial list of profiles ready for collection, we used Twitter Stream API and the geographic filter to ensure that the collected Tweets are within the specified region. Moreover, we collected the Twitter metadata for each user such as characteristics of the Twitter profile (that are independent of tweet content), to determine demographic information.

---

[2] http://pan.webis.de/clef17/pan17-web/author-profiling.html
[3] https://github.com/tweepy/tweepy

As the data collected from social media is usually noisy, we wrote a script to clean the collected Tweets from non-textual content such as images and URLs. Moreover, we filtered all non-Arabic content from the collected data.

For each region, we collected the profiles of 100 users with at least 2000 posted Tweets. For all users, we downloaded up to their last 3240 tweets, which is the limit imposed by Twitter API.

We ended up with a minimum of 200K Tweets per region and a total of 2.4 Million Tweets corpus (Zaghouani and Charfi 2018a).

During the data collection process, we tried to expand our coverage as much as possible taking into consideration the resources and the budget available, we were able to collect a balanced Tweets corpus from 11 Arabic regions representing a total of 16 countries from a total of 22 Arabic countries members of the Arab league as shown in Table 2. We tried to select the data as randomly as possible by avoiding well-known/famous and influential users.

| Dialect | Region |
|---|---|
| Moroccan | 1. Morocco |
| Algerian | 2. Algeria |
| Tunisian | 3. Tunisia |
| Libyan | 4. Libya |
| Egyptian | 5. Egypt |
| Sudanese | 6. Sudan |
| Lebanese | 7. North Levant |
| Syrian | 7. North Levant |
| Jordanian | 8. South Levant |
| Palestinian | 8. South Levant |
| Iraqi | 9. Iraq |
| Qatari | 10. Gulf |
| Kuwaiti | 10. Gulf |
| Emirati | 10. Gulf |
| Saudi | 10. Gulf |
| Yemeni | 11. Yemen |

Table 2: Dialects and regions selected in the corpus

Once our data is ready, we started a manual annotation step for the collected user profiles in order to: (a) validate the data collected; (b) annotate each user with the age and gender; (c) confirm the dialect used by the users and check if she is a native or non-native speaker of Arabic.

We created general and specific annotation guidelines and we employed a group of annotators to perform the manual annotation for each annotation task.

## 4. Annotation Guidelines

The annotation guidelines usually document the core of the annotation policy in any given corpus annotation project. Our guidelines are tailored to each of the four annotation tasks within the context of our project: the gender, the age, the dialect and whether the user is a native Arabic speaker or not.

We describe the process of how to annotate each of these tasks, including how to deal with borderline cases. We provided many annotated examples based on our guidelines to illustrate the annotation rules and exceptions for each task. We adopted an iterative approach to develop our guidelines, which includes many revisions and updates as needed in order to reach a consistent set of instructions. For instance, several changes to the guidelines were needed to address the issue of age identification task due to the complexity and the difficulty of this particular task.

The annotations were done by carefully analyzing each of the user's profiles, their tweets, and when possible, we instructed the annotators to use external resources such as personal web pages or blogs as well as other social networks such as LinkedIn and Facebook. We created profiles validation guidelines and task-specific guidelines to annotate the users.

### 4.1 Profiles Validation Guidelines

To ensure the suitability of the corpus collected for the author profiling task we wrote the annotation guidelines. Moreover, we clearly instructed the annotators on how to validate or exclude the collected Twitter profiles from our data. Finally, we set simple and clear rules and requirements as listed below:

- The profile should be public as we cannot retrieve the data from private or protected profiles.
- The tweets should have been mostly written in the given regional dialect. Moreover, the Tweets should not be mostly written in standard Arabic or any other language such as English or French.
- The profile should represent an actual person (i.e., not a company).
- The profiles posting lots of images and using applications to automatically post daily messages by bots are also filtered out.

### 4.2 Gender Annotation Guidelines

For some accounts, the annotators were not able to identify the gender as this was based in most of the cases on the name of the person or his/her profile photo and in some cases by their biography or profile description. In case this information is not available, we instructed the annotators to read the user posts and find linguistic indicators of the user's gender.

Like many other languages, Arabic conjugates verbs through numerous prefixes and suffixes and the gender is sometimes clearly marked such as in the case of the verbs ending in taa marbuTa (ة/ـة) which is usually of feminine gender as shown in the example in Table 3.

| Form | Sentence |
|---|---|
| English masc. / fem. Form | *I am thirsty* |
| Arabic masc. form | أنا عطشان /ana Atshaan/ I am thirsty (Masc.) |
| Arabic fem. Form | أنا عطشان**ة** /ana atshaan**a**/ **I am thirsty (fem.)** |

Table 3: Taa Marbuta gender marker in the Arabic verbs

### 4.3 Age Annotation Guidelines

In order to annotate the users for their age, we used three categories: under 25 years, between 25 years and 34 years, and 35 years and above.

In our guidelines, we asked the annotators to check if the user birth year is available in their Twitter profile. Depending on the dialect region, 4 to 7 % of the users put this information in their public profile. We also asked the annotators to read the latest 100 tweets of the user first for validating their dialect and second for finding any age-related hints. For example, some users had tweets such as "I just turned 25". In some cases, the annotators found some hints indicating that the users were high school or university students such as tweets about exams, schools, and university breaks, etc.

Next, we asked the annotators to retrieve the full name of the user also from their profile and when available search for that name on search engines as well as on other social networks such as LinkedIn and Facebook. The search retrieved for some users their personal homepage or their blog, which could contain their age information. As some Twitter users put their photo in their profile picture this helped the annotators in match twitter users with their respective web page, blog, or social media account on Facebook, Instagram, and LinkedIn. Also other information from the Twitter profile such as the name of the city they live in as well as their job description was helpful for matching accounts on different social networks. In the case of LinkedIn, the graduation year from school or university and also the professional experience were helpful in determining the age group. For example, someone who graduated from university in the year 2000 is certainly above 35 years.

In the last step, if a Twitter profile photo is available the annotators were asked to estimate the age based on that photo (as well as any other photos that the same person may have on their Twitter account in the photos section). Then, we instructed the annotators to use the artificial intelligence based Microsft service How-Old.Net[4], which takes an image in input and determines the subject's age and gender.

[4] https://how-old.net

as shown in Figure. 1. In addition, we wrote a program that automatically retrieves the profiles photos for all selected users and retrieves their age and gender using Microsoft Face API[5]. Even though both tools are from Microsoft they delivered slightly different results.

In the cases, in which age estimation was not possible we replaced the respective Twitter accounts by others of the same gender and from the same region. The newly added accounts were selected so that they provide indications and hints about the age as explained above.

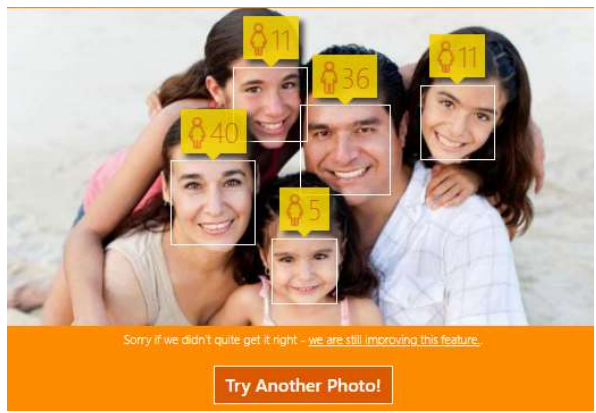

Figure 1: Automatic age estimate sample by How-Old.Net. Photo credit: GSCSNJ (Creative Commons)

### 4.4 Dialect variety Annotation Guidelines

As the dialect and the regions are known in advance to the annotators, we instructed them to double check and mark the cases in which the user appears to be from a different dialect group. This is possible despite our initial filtering based on distinctive regional keywords. We noticed that in more than 90% the profiles selected belong to the specified dialect group. For the 10% remaining, we observed many cases of people borrowing terms and expressions from other dialects such as in the case of the word بزاف Bizzef 'many' which is typically used in Algerian dialect and also in the Moroccan dialect. In case of doubt, the annotators were instructed to use Google search to check the usage frequency of a given word and to which dialect it is mostly associated.

### 4.5 Native Language Annotation Guidelines

The goal of this annotation task is to mark and identify Twitter profiles with a native language other than Arabic, so they are considered as Arabic L2 speakers. In order to help the annotators identify those, we instructed them to look for the following cues in order to identify the non-native Arabic users:

- Essays produced by learners of Arabic as second language differ from those of natives, not only quantitatively but also qualitatively. Their writings display very different frequencies of words, phrases, and structures, with some items overused and others significantly underused.
- Sentences written by Arabic L2 speaker have often a different structure and are not as fluent as sentences produced by a native speaker even when no clear mistakes can be found.
- Style: Arabic L2 Tweets texts may be written in a style that is unfamiliar or unnatural to native speakers although the word order is acceptable, and the sentence conveys the meaning correctly.

[5] https://azure.microsoft.com/en-us/services/cognitive-services/face/

Non-native Tweets also contain varying degrees of grammatical, orthographic and lexical errors generally not produced by native speakers. When identifying non-native users, we instructed the annotators to focus on lexical choice errors and syntactic errors as detailed below:

- Word Choice Errors: These include the obvious use of an incorrect word in a given context. Word choice errors are particularly frequent in the L2 Arabic student essays.
- Syntactic Errors: These include a wrong agreement in gender, number, definiteness or case as well as wrong case assignment, wrong tense use, wrong word order.

## 5. Annotation Logistics

The annotation of a large scale corpus requires the involvement of a team of annotators. In our project, the annotation effort was led by a lead annotation manager who is responsible for the whole annotation task. This includes compiling the data, the annotation of the gold standard Inter-Annotator Agreement (IAA) portion of the corpus, writing the annotation guidelines, hiring and training the annotators, evaluating the quality of the annotation, monitoring and reporting on the annotation progress. To ensure the suitability of the annotators for the various annotation tasks, we selected university level annotators with a good knowledge of the Arabic regional dialects selected. Furthermore, the annotators were screened by doing a limited number of annotation tasks, once hired, they spent a training period of two weeks. During the training period, the annotators read the guidelines, held several group meetings and completed some tasks before starting the official annotation phase.

During the annotation phase and to ensure the quality of the annotated corpus, the annotation manager assigned files to be done by all the annotators and later on, their annotation was compared to compute their Inter-Annotator agreement scores (IAA). Furthermore, a communication message board was provided as space for the annotators to post their questions, add comments, report issues and get feedback from the annotation manager as well as the other annotators. We encouraged the annotators to use this way of communication in order to keep track of all the issues faced and to have an interaction archive that can be used later on to improve the current version of the guidelines.

## 6. Evaluation

We evaluate the Inter-annotator agreement (IAA) to quantify the extent to which independent annotators, excluding the lead annotator, trained using our guidelines, agree on the annotations. A high level of agreement between the annotators indicates that the annotations are reliable and the guidelines are useful in producing homogeneous and consistent dataset. We created a gold standard dataset of 110 Twitter profiles representing the 11 regions to evaluate the annotators and their application of the guidelines.

During the evaluation, we assigned in a blind way, the sample dataset to all the annotators without any mention to them, so that it was considered as a regular annotation exercise from their end. Later on, we measured the Inter-annotator agreement using Cohen's kappa formula. At the end of the evaluation, we computed the average Kappa scores obtained by the annotators and listed in Table 4. For the gender annotation, they obtained a high score of 0.95, for the age annotation an average score of 0.80, for the dialect identification a score of 0.92 and finally for the native language annotation a relatively low score of 0.70.

As observed, the gender annotation task score was the highest with a near perfect agreement of 95%. For the dialect identification task, some annotators were confused by a few similarities that exists between some dialects such as the Moroccan dialect and the Algerian dialect and also by the Qatari dialect and some other Gulf dialects.

The age identification task proved to be a difficult task, especially with the absence of clear cues and indicators such the birth year, graduation year and the absence of a profile photo.

Finally, the native language identification ranked last as it could be very hard to find due to the lack of cues. Overall, we believe that the annotation agreement is above the acceptable range given the difficulty of the tasks.

| Task | Kappa Score |
|---|---|
| Gender Annotation | 0.95 |
| Dialect Annotation | 0.92 |
| Age Annotation | 0.80 |
| Native Language | 0.70 |

Table 4: Inter-annotator agreement in terms of average Kappa score; the higher the better

## 7. Conclusion

We presented a set of guidelines and our annotation pipeline to build a large 2.4M annotated Tweets Arabic author profiling corpus called ARAP-Tweet. We summarized our general and dialect-specific guidelines for each of the 11 Arabic dialectal regions collected. The guidelines and the resource created could be used for tasks other than author profiling. In the future, we plan to release the guidelines and the corpus[6] to the research community during the 3rd Workshop on Open-Source Arabic Corpora and Processing Tools.[7] Moreover, the corpus will be provided to the participants of the author profiling task during the 18th evaluation lab on digital text forensics, PAN @ CLEF 2018.[8]

---

[6] As per the Twitter agreement and policy and in order to protect the privacy of the users, we will only distribute the Tweet IDs in the public data release.

## 9. Acknowledgements

This publication was made possible by NPRP grant 9-175-1-033 from the Qatar National Research Fund (a member of Qatar Foundation). The findings achieved herein are solely the responsibility of the authors.

## 8. Bibliographical References

## 9. Language Resource References

---

[7] http://edinburghnlp.inf.ed.ac.uk/workshops/OSACT3/

[8] http://pan.webis.de/clef18/pan18-web/index.html